\documentclass[sigconf]{acmart}

\usepackage{booktabs} 
\usepackage{amsmath}
\usepackage{bm}
\usepackage{algorithm}
\usepackage{algorithmic} 

\DeclareMathOperator*{\argmax}{argmax}
\usepackage{multicol,multirow}
\usepackage{enumitem}

\copyrightyear{2022}
\acmYear{2022}
\setcopyright{acmlicensed}\acmConference[SAC '22]{The 37th ACM/SIGAPP Symposium on Applied Computing}{April 25--29, 2022}{Virtual Event, }
\acmBooktitle{The 37th ACM/SIGAPP Symposium on Applied Computing (SAC '22), April 25--29, 2022, Virtual Event, }
\acmPrice{15.00}
\acmDOI{10.1145/3477314.3507319}
\acmISBN{978-1-4503-8713-2/22/04}

\begin{document}
\title{EiFFFeL: Enforcing Fairness in Forests by Flipping Leaves}
\author{Seyum Assefa Abebe}
\affiliation{  
\institution{Ca’ Foscari University of Venice}
  \city{Venice} 
  \state{Italy} 
}
\email{seyumassefa.abebe@unive.it}

\author{Claudio Lucchese}
\affiliation{  
\institution{Ca’ Foscari University of Venice}
  \city{Venice} 
  \state{Italy} 
}
\email{claudio.lucchese@unive.it}

\author{Salvatore Orlando}
\affiliation{  
\institution{Ca’ Foscari University of Venice}
  \city{Venice} 
  \state{Italy} 
}
\email{orlando@unive.it}

\renewcommand{\shortauthors}{Abebe et al.}

\begin{abstract}
Nowadays Machine Learning (ML) techniques are extensively adopted in many socially sensitive systems, thus requiring to carefully study the fairness of the decisions taken by such systems. Many approaches have been proposed to address and to make sure there is no bias against individuals or specific groups which might originally come from biased training datasets or algorithm design. In this regard, we propose a fairness enforcing approach called EiFFFeL --Enforcing Fairness in Forests by Flipping Leaves-- which exploits tree-based or leaf-based post-processing strategies to relabel leaves of selected decision trees of a given forest.
Experimental results show that our approach achieves a user-defined group fairness degree without losing a significant amount of accuracy. 
\end{abstract}

\ccsdesc[500]{Computing methodologies~Machine learning algorithms}
\ccsdesc[500]{Human-centered computing}

\keywords{Fair machine learning, Random forest, Group Fairness}

\maketitle

\section{Introduction}

Machine Learning is used in a wide range of systems, such as health care \cite{de2018clinically,kourou2015machine}, education \cite{oneto2017dropout,oneto2019learning,papamitsiou2014learning}, financial lending \cite{byanjankar2015predicting,malekipirbazari2015risk}, and social services \cite{now2018litigating,eubanks2018automating}, to facilitate decision making and automate services which has a critical implications to individuals and communities. This extensive use of machine learning creates a growing concern, as algorithms might introduce far-reaching bias that treats individuals or groups unfairly, based on certain characteristics such as age, race, gender, or political affiliation.  Thus, it is becoming very important to develop fairness aware algorithms.

In recent years many methods have been developed addressing both individual-based and group-based fairness. Most of the works tackles issues of discovering discrimination, and adding solutions to eliminate such discrimination to have fair and accurate decisions \cite{zhang2019fairness,zliobaite2015survey}. The bias mitigation approaches on either the training data or the learned model can be categorized into three main groups \cite{d2017conscientious}: $(1)$ Pre-processing approaches such as reweighting \cite{calders2009building}, massaging \cite{kamiran2009classifying}, aiming to eliminate discrimination at the data level; $(2)$ In-processing algorithms such as \cite{bhaskaruni2019improving,kamiran2010discrimination} addressing discrimination by modifying algorithms; $(3)$ post-processing methods such as \cite{hajian2015discrimination,kamiran2010discrimination} take the learned model and change the predicted labels appropriately to meet fairness requirements.  

Algorithms proposed recently in bias mitigation has focused on neural networks. However, the efficiency and explainability of tree ensembles for many applications makes them preferable to be implemented in many areas. Even though there are few works focused on studying fairness for trees and tree ensembles, notably \cite{kamiran2010discrimination,raff2018fair,zhangfaht,grari2020achieving}, most of them are focused on single decision tree classifiers and in-processing approaches. Our interest mainly lies in developing fair random forest classifiers with post-processing approaches designed to relabel leaves with accuracy and discrimination constraints. We take advantage of implementing a post-process approach, in which we do not require to know the training process.       

{\bf Contributions.}
We focus on decision tree ensembles for binary classification tasks susceptible to group discrimination 
with respect to attributes sensitive classes such as age, gender, race, etc.
We propose a post-processing approach named EiFFFeL --Enforcing Fairness in Forests by Flipping Leaves-- that given a forest,
however trained, selects a subset of its leaves and changes their predictions so as to reduce the discrimination degree of the forest.

We summarize the main contributions of our work as follows.
\begin{enumerate}[noitemsep,topsep=4pt]
    \item  We propose an iterative leaf flipping post-processing algorithm to ensure group fairness .
    \item We devise  tree-based and leaf-based flipping methodologies on top of random forest classifier to enforce fairness.
    \item We report experimental evaluations of group fairness on three different datasets, aiming to empirically show the effectiveness of our method.
\end{enumerate}

\section{Fairness in Machine Learning}

Without loss of generality, we consider a binary classifier $g:\mathcal{X \to Y}$ that maps an input feature vector $\bm{x} \in \mathcal{X}$ to a binary class label $y \in \mathcal{Y} = \{0,1\}$.
Among the attributes in the feature space $\mathcal{X}$,
a binary attribute called \textit{sensitive feature} ${S} \in \{0,1\}$ identifies the aspects of data which are socio-culturally precarious for the application of machine learning. Specifically, given $x \in \mathcal{X}$ and $x.S$ the value of the sensitive attribute $S$ for the given instance, if $x.S=0$ then we say that $x$ belongs to the \textit{unprivileged group} that could possibly be discriminated.

\subsection{Fairness and Discrimination Definitions}
To achieve non-discriminatory and  fair machine learning model, it is essential to first define \textit{fairness}. In a broad context, fairness can be seen from an individual or a group point of view. \textit{Individual fairness} requires that similar individuals being treated similarly. 
\textit{Group fairness} requires fairness of the classification model to apply on the two groups, defined through the binary sensitive feature $S$ \cite{dwork2012fairness}. Our work focuses on group fairness, in which a group of individuals identified by $S$ risks for experiencing discrimination. 

We define the discrimination of a classifier measured by \textit{group fairness} as follows.
Recall that attribute $S=0$ identifies the unprivileged group, while $S=1$ corresponds to the  privileged one, whose members are not discriminated but rather favoured by a learnt ML model.  
Moreover, we assume that the values $1$ and $0$ of class label $Y$ represent \textit{favorable} and \textit{unfavorable} outcomes, respectively. For example, $Y=1$ might correspond to the decision of granting a  loan, thus favouring a bank customer.

A classifier $g$ applied over $x \in \mathcal{X}$ is non-discriminatory if its prediction $g(\bm{x})$ is statistically independent of the sensitive attribute $S$. Hence, a classifier is fair if both groups have equal probability of being classified as belonging to the favorable class, which is the desirable outcome. 

Using the problem formalization by \cite{kamiran2010discrimination}, the discrimination of a model $g$ with respect to a sensitive attribute $S$ and a dataset $\mathcal{D}=\{(\bm{x}_i, y_i)\}_{i=1}^N$ can be computed as follows:

\begin{multline*}
    disc_{\mathcal{D},S,g} := \frac{|\{ (\bm{x},y) \in \mathcal{D} ~|~ \bm{x}.S=1 \wedge g(\bm{x})=1\}|}{|\{(\bm{x},y) \in \mathcal{D}~|~ \bm{x}.S=1\}|}\\ - \frac{|\{(\bm{x},y) \in \mathcal{D}~|~ \bm{x}.S=0 \wedge g(\bm{x})=1\}|}{|\{ (\bm{x},y) \in \mathcal{D} ~|~ \bm{x}.S=0 \}|} ,
\end{multline*}

\noindent where $\bm{x}.S$ refers to the sensitive attribute of the instance $\bm{x}$. When $S$ and $\mathcal{D}$ are clear from the context we simply use the notation $disc_g$.

To clarify the above definition, let's consider the case of a classifier $g$ used by the HR staff of a company. The classifier $g$ suggests hiring when $g(\bm{x})=1$ vs.\ not hiring when $g(\bm{x})=0$. We may wonder whether the classifier favours \textit{men} ($S=1$) over \textit{women} ($S=0$). The value of $disc_g$ is large if the ratio of men with a favorable hiring prediction is larger than the ratio of women with a favorable hiring prediction. 
By minimizing $disc_g$ we can provide a fairer classifier w.r.t.\ the gender attribute.

 \subsection{Related Works}
Notably, in recent years works  identifying and solving bias in machine learning algorithms have progressed. \textit{Pre-processing}, \textit{in-processing}, and \textit{post-processing} approaches have been used to mitigate and quantify bias coming from training data, learning algorithms, or the interaction between the twos. 

Algorithms which are identified in the \textit{Pre-processing} category deal with discrimination at the dataset level by altering its distribution to ensure there is no bias against a specific group or individual. This can be achieved by removing the sensitive attribute, re-sampling the data, or changing class labels. One of the well known pre-processing method is massaging \cite{kamiran2009classifying}, which changes the class labels of a subset of carefully selected instances. Another work in this category is re-weighting \cite{calders2009building}, which assigns different weights to different groups of the dataset to reduce bias. A re-sampling approach in \cite{calmon2017optimized} limits the sample size to control discrimination. 

In the \textit{In-processing} bias mitigation algorithms, discrimination is accounted during the training phase of the learning algorithm. Strategies in this group take different approaches to discount discrimination by including fairness penalty into the loss function such as in \cite{zafar2017fairness}, which integrates decision boundary covariance constraint for logistic regression. In \cite{aghaei2019learning} regularization terms are added to penalize discrimination  in mixed-integer optimization framework of decision tree. Another interesting work is \cite{calders2010three}, which proposes three approaches for fairness-aware Na\"ive Bayes classifiers. The approaches are: altering the decision distribution until there is no more discrimination, building a separate model for each sensitive group to remove the correlation between sensitive attribute and class label, and adding latent variable representing unbiased label.

Kamiran \textit{et al.} \cite{kamiran2010discrimination} included a discrimination factor into the information gain splitting criterion of a single decision tree classifier by considering the split of a node under the influence of a sensitive feature, i.e., before a node split happens  not only the usual purity w.r.t.\ to the target label is calculated, but also the purity of the split w.r.t.\ the sensitive feature. Three alternative splitting criteria are given based on the way discrimination is accounted. The first option is subtracting discrimination gain from accuracy gain, which allows for a split if it is non-discriminatory, second option is an accuracy-discrimination trade-off split where the accuracy gain is divided by discrimination gain to have the final gain value. The third option is adding the accuracy and discrimination gain to decide the best feature to split a node. The authors claim the additive information gain criterion produces a lower discrimination. We also implement this method for the base trees of our forest and evaluate the impact of it to the overall forest discrimination value.
Finally, authors propose an additional relabeling of some leaves of the tree so as to further reduce its discrimination degree.

A recent work, called Distributed Fair Random Forest (DFRF) \cite{fantin2020distributed} exploits randomly generated decision trees and filters them by their fairness before adding them to the forest.
This is achieved through a hyper-parameter fairness constraint, which forces to accept only decision trees with statistical parity below the given threshold. The generation and fairness thresholding of each individual tree can be done in distributed framework that optimizes the trade-of between discrimination and accuracy of the tree before being added to the forest. Furthermore, this algorithm uses randomness constraint to train base trees in which one feature is randomly selected to split a node for building a randomized decision tree.   

\textit{Post-processing} mitigation approaches focus on adjusting the final output of the trained model rather than the underline loss function or training data. 
The algorithms discussed in \cite{hardt2016equality,pleiss2017fairness}  aim at achieving same error rates between privileged and unprivileged groups,  
  
\cite{hardt2016equality} uses equalized odd and equalized opportunity to promote features which are more dependent on the target label than the sensitive attribute. While 
in \cite{pleiss2017fairness} the proposed algorithm aims to achieve both privileged and unprivileged groups to have the same false negative rate and false positive rate 
by taking into account a calibrated probability estimates.
Another post-processing algorithm called Reject Option based Classification (ROC) \cite{kamiran2012decision} takes in to consideration the decision boundary of  classifiers;  in a region where uncertainty is high, it gives favorable outcomes to the unprivileged group and unfavorable outcomes to the privileged group to reduce discrimination. 

Among the various works, the closest to our proposal is  \cite{kamiran2010discrimination}, which relabels leaves of a single tree classifier with a small effect on the model accuracy.
We borrow from this approach and propose a novel algorithm for enforcing fairness in forests of decision trees.

 \begin{table}[H]
    \centering
    \footnotesize
    \caption{\label{tab:notation} Notation Summary}
    \begin{tabular}{@{}c|p{.6\columnwidth}@{}}
    \toprule
        Symbol & Meaning \\ 
    \midrule
        $\mathcal{D}$ &  Dataset \\
        $S$  & Sensitive feature\\
        $\lambda$ &  leaf  \\
        $\Lambda$ & Set of Leaves to be flipped\\
        $disc_{ \mathcal{F}}$ & Forest discrimination\\
        $disc_{ \mathcal{T}}$ & Tree discrimination\\
        $accu_{ \mathcal{F}}$ & Forest accuracy \\
        $\Delta disc_{ \mathcal{\lambda}}$ & change in discrimination after flipping\\
        $\Delta accu_{ \mathcal{\lambda}}$ & change on accuracy after flipping \\
        $\delta$ & Ratio of change in accuracy and discrimination \\
    \bottomrule
    \end{tabular}
\end{table}

\section{The EiFFFel Algorithm}

We propose a novel post-processing algorithm named  EiFFFeL that, given a forest of decision trees for a binary classification task, modifies the prediction of a carefully chosen set of leaves so as to reduce the forest's discrimination degree.
This process is named leaf relabeling, or, since we are focusing on a binary prediction task, \emph{leaf flipping}. 

The rationale is to flip the prediction of the leaves that contribute the most to the model discrimination degree so as to make them fair.
Recall that the score  $disc_{\mathcal{D},S,g}$ adopted to evaluate the model's discrimination depends on the number of privileged/unprivileged instances with a favorable prediction.
Therefore, by flipping a leaf label we can increase or decrease the number of instances that contribute to the discrimination score.
Note that, while leaf relabeling can be done judiciously so as to reduce discrimination, modifying the leaf predictions determined at training time may reduce the accuracy of the whole forest.

Therefore the goal of EiFFFeL is to find a sweet-spot in the accuracy vs.\ discrimination trade-off.
While leaf relabeling was introduced by \cite{kamiran2010discrimination} for a single tree, we improve such strategy and extend it to a forest of decision trees. 

In this work we focus on Random Forests ensembles, which, for their high accuracy and limited bias, are an optimal candidate for building a fair classifier. The approach is however general and we leave to future work the application to other tree ensembles, such as those obtained by bagging and boosting approaches.

The proposed EiFFFel algorithm accepts a user-defined maximum discrimination constraint $\epsilon$ and a minimum relative accuracy drop constraint $\alpha$. 
Given a forest $\mathcal{F}$, it iteratively modifies the prediction associated with a subset of the leaves of $\mathcal{F}$, until either the desired discrimination $\epsilon$ is achieved, or the maximum required accuracy drop $\alpha$ is hit.

Below we first illustrate the \emph{Leaf Scoring} strategy used to find the most discriminative leaves of a tree, and then we illustrate two variants of the EiFFFeL algorithm.

\begin{algorithm}[t]
	\renewcommand{\algorithmicrequire}{\textbf{Input:}}
	\renewcommand{\algorithmicensure}{\textbf{Output:}}
	\caption{\sc Score\_Leaves}
	\label{alg:getL}
    \begin{algorithmic}[1]
    \small
    \REQUIRE Decision Tree $\mathcal{T}$ \\  Dataset $\mathcal{D}$ \\ Sensitive feature $S$ 
    \ENSURE Candidate flipping leaves $\Lambda$ 
     \STATE $\Lambda \gets \emptyset$
    \FORALL{$\lambda \in  \mathcal{T} ~|~ \neg \lambda.flipped$}

    \STATE $\Delta accu_{ \lambda}\gets -abs\left( \frac{ | \mathcal{D}^\lambda_{y= 1} | - |\mathcal{D}^\lambda_{y= 0 }| }{|\mathcal{D}|} \right)$  \label{lst:line:line4}
    
    \STATE $\Delta disc_{\lambda} \gets sign\left(| \mathcal{D}^\lambda_{y=1} | - |\mathcal{D}^\lambda_{y=0} |\right)\cdot\left( \frac{ |\mathcal{D}^\lambda_{S=1} |}{|\mathcal{D}_{S=1} |}
    -  \frac{|\mathcal{D}^\lambda_{S=0}|}{|\mathcal{D}_{S=0}|} \right) $ \label{lst:line:line3}

    \STATE $\delta \gets \frac{\Delta disc_{\lambda}}{\Delta accu_{ \lambda}}$ \label{lst:line:line5}
    \IF{$\delta \geq 0$}
      \STATE $\lambda.score \gets \delta$
      \STATE $\Lambda \gets \Lambda \cup \{\lambda\}$
    \ENDIF
    \ENDFOR
    \RETURN $\Lambda $
    \end{algorithmic}
\end{algorithm}

\subsection{Leaf Scoring}

EiFFFeL borrows from \cite{kamiran2010discrimination} a simple strategy for scoring leaves according to their impact $\Delta accu_{ \lambda}$ and $\Delta disc_{ \lambda}$ on accuracy and discrimination respectively. Then, the ratio $\delta$ between the two is used as a score to greedily select the best leaves to be \emph{flipped}.

We proceed as described in Alg.~\ref{alg:getL}.
We consider only leaves of the tree that were not flipped during previous iteration of the EiFFFeL algorithm (see subsection below). For those leaves we compute the accuracy and discrimination variation in the case of flipping the leaf prediction. We illustrate shortly the computations below, please refer to \cite{kamiran2010discrimination} for a more detailed description.

The change in accuracy $\Delta accu_{ \lambda}$ clearly depends on the number of instances of $\mathcal{D}$ that fall into the leaf $\lambda$ denoted with $\mathcal{D}^\lambda$. The training process sets the leaf prediction to the majority class among such instances. Therefore, when flipping the leaf prediction the accuracy may only decrease depending on the instances with label $1$ and $0$, denoted by $\mathcal{D}^\lambda_{y=1}$ and $\mathcal{D}^\lambda_{y=0}$ respectively. The difference between the size of these two sets results in the accuracy loss as computed in line~\ref{lst:line:line4}.

The change in discrimination $\Delta disc_{ \lambda}$ depends on the number of privileged and unpriviledged instances that fall in the leaf $\lambda$ respectively denoted by $\mathcal{D}^\lambda_{S=1}$ and $\mathcal{D}^\lambda_{S=0}$, and on their analogous on the whole dataset $\mathcal{D}_{S=1}$ and $\mathcal{D}_{S=0}$. If the leaf prediction equals $1$ (favourable class), then increasing $\mathcal{D}^\lambda_{S=1}$ would increase the discrimination, while increasing $\mathcal{D}^\lambda_{S=0}$ would decrease it. The opposite holds if the prediction of the leaf equals $0$ (unfavourable class). As the original leaf prediction depends on the majority of the instances between $\mathcal{D}^\lambda_{y=1}$ and $\mathcal{D}^\lambda_{y=0}$, the sign of their difference is used to correct the above contributions as computed in line~\ref{lst:line:line3}.

The ratio $\delta=\Delta disc_{ \lambda}/\Delta accu_{ \lambda}$ is positive if the flipping generates a discrimination drop, and it is large if the benefit to discrimination is larger than the harm to accuracy. If the value of $\delta$ is positive, then this is stored with the leaf $\lambda$, and $\lambda$ is recorded into the set of candidate leaves $\Lambda$. The set $\Lambda$ is eventually returned and exploited during the iterations of EiFFFeL.

\subsection{EiFFFeL Leaf Flipping Strategies}

By exploiting the scoring technique discussed before, we propose two strategies to choose which trees and which leaves in those trees to flip. 

The first strategy, named \textit{Tree-based Flipping}, is illustrated in Alg.~\ref{alg:mlr}.
During each iteration of EiFFFeL, the tree $\mathcal{T}^\dagger$ with the largest discrimination degree is greedily selected: this is the best tree to be attacked in order to significantly reduce the discrimination of the full forest. Then, we use the previous scoring technique to find the set of leaves $\Lambda$ in $\mathcal{T}^\dagger$ that should be relabeled. If $\Lambda$ is not empty, the predictions $\lambda.pred$ of such leaves will be flipped. Then, the whole tree is marked as already flipped. The selection is repeated by considering only the remaining non-flipped trees. The algorithm ends when all trees have been flipped, or when the desired discrimination $\epsilon$ is achieved, or when tolerated accuracy drop $\alpha$ is met. Note that the accuracy drop is computed by comparing the accuracy of the original forest with the accuracy of the current forest after the flipping step.

\begin{algorithm}[t]
    \small
	\renewcommand{\algorithmicrequire}{\textbf{Input:}}
	\renewcommand{\algorithmicensure}{\textbf{Output:}}
	\caption{EiFFFeL-TF (Tree-based Flipping)}
	\label{alg:mlr}
	\begin{algorithmic}[1]
	\REQUIRE Random Forest classifier $\mathcal{F}$   \\ Discrimination Constraint $\epsilon \in [0,1]$ \\ Accuracy Constraint $\alpha \in [0,1]$ \\  Training Dataset $\mathcal{D}$ \\ Sensitive feature $S$ 
	\ENSURE Fair Random Forest $\mathcal{F}$ 

    \FORALL{$\mathcal{T} \in \mathcal{F}$}
        \STATE{$\mathcal{T}.flipped \gets false$}
        \FORALL{$\lambda \in \mathcal{T}$}
            \STATE{$\lambda.flipped \gets false$}
        \ENDFOR
    \ENDFOR

  \STATE $accu_{ \mathcal{F}}^*\gets  \frac{ | \mathcal{D}_{y= 1} \wedge \mathcal{F}(\bm{x})=1  |~+~ |\mathcal{D}_{y= 0 }\wedge \mathcal{F}(\bm{x})=0| }{|\mathcal{D}|} $
  
  \STATE{$\Delta accu_{\mathcal{F}} \gets 0$}
  
    \WHILE{$|\{\mathcal{T} \in \mathcal{F} ~|~ \neg\mathcal{T}.flipped\}|>0 \wedge\hfill \hspace*{\fill}\linebreak
    \hspace*{3em}\ disc_{\mathcal{D},S,\mathcal{F}} > \epsilon\ \wedge\ \Delta accu_{\mathcal{F}} < \alpha$}
    
	\STATE $\mathcal{T}^\dagger \gets \argmax_{\mathcal{T} \in \mathcal{F}}\ disc_{\mathcal{D},S,\mathcal{T}}$

	\STATE $\Lambda \gets$ {\sc Score\_Leaves($\mathcal{T}^\dagger,\mathcal{D},S$)} 
	\IF{$\Lambda\neq\emptyset$}
   \FORALL{$\lambda \in \Lambda$}
        \STATE{$\lambda.pred = 1 - \lambda.pred$}
   \ENDFOR
   \ENDIF
   \STATE{$\mathcal{T}^\dagger.flipped = true$}

\STATE{$accu_{ \mathcal{F}}\gets  \frac{ | \mathcal{D}_{y= 1} \wedge \mathcal{F}(\bm{x})=1  |~+~ |\mathcal{D}_{y= 0 }\wedge \mathcal{F}(\bm{x})=0| }{|\mathcal{D}|} $}
  \STATE{$\Delta accu_{\mathcal{F}} \gets accu_{ \mathcal{F}}^* - accu_{ \mathcal{F}}$}
	\ENDWHILE

	\RETURN $\mathcal{F}$
	\end{algorithmic}
\end{algorithm}

Such tree-based strategy might be too  aggressive, as it immediately flips all the candidate leaves of the selected tree. Indeed, only a few leaves may be  sufficient to meet our discrimination and accuracy requirements. 
Therefore we propose a second strategy, named  \textit{Leaf-Based Flipping}, illustrated in Alg~\ref{alg:slr}. As in the former strategy, we first select  the tree $\mathcal{T}^\dagger$ with the largest discrimination. Then we use the leaf scoring technique to find a set of candidate leaves from $\mathcal{T}^\dagger$. If such set is empty, e.g., because they were already flipped or they cannot improve the discrimination, the full tree is marked as flipped and the procedure is repeated on the remaining non-flipped trees. Otherwise, the leaf with the largest score $\lambda^\dagger$ is selected, marked as flipped, while its prediction is inverted. The process is repeated until all trees have been flipped, or the desired discrimination $\epsilon$ is achieved, or the tolerated accuracy drop $\alpha$ is met. 

We argued that the Leaf-based approach exploits a more fine-grained tuning of the given forest, and therefore it can achieve the desired accuracy with a smaller set of alterations. Indeed, reducing the flips applied to the forest provides a larger accuracy.

\begin{algorithm}[t]
    \small
	\renewcommand{\algorithmicrequire}{\textbf{Input:}}
	\renewcommand{\algorithmicensure}{\textbf{Output:}}
	\caption{EiFFFeL-LF (Leaf-based Flipping)}
	\algsetup{linenosize=\small}
	\label{alg:slr}
	\begin{algorithmic}[1]
	\REQUIRE Random Forest classifier $\mathcal{F}$   \\ Discrimination Constraint $\epsilon \in [0,1]$ \\ Accuracy Constraint $\alpha \in [0,1]$ \\  Training Dataset $\mathcal{D}$ \\ Sensitive feature $S$ 
	\ENSURE Fair Random Forest $\mathcal{F}$ 

    \FORALL{$\mathcal{T} \in \mathcal{F}$}
        \STATE{$\mathcal{T}.flipped \gets false$}
        \FORALL{$\lambda \in \mathcal{T}$}
            \STATE{$\lambda.flipped \gets false$}
        \ENDFOR
    \ENDFOR

  \STATE $accu_{ \mathcal{F}}^*\gets  \frac{ | \mathcal{D}_{y= 1} \wedge \mathcal{F}(\bm{x})=1  |~+~ |\mathcal{D}_{y= 0 }\wedge \mathcal{F}(\bm{x})=0| }{|\mathcal{D}|} $
  
  \STATE{$\Delta accu_{\mathcal{F}} \gets 0$}
  
    \WHILE{$|\{\mathcal{T} \in \mathcal{F} ~|~ \neg\mathcal{T}.flipped\}|>0 \wedge\hfill \hspace*{\fill}\linebreak
    \hspace*{3em}\ disc_{\mathcal{D},S,\mathcal{F}} > \epsilon\ \wedge\ \Delta accu_{\mathcal{F}} < \alpha$}

	\STATE $\mathcal{T}^\dagger \gets \argmax_{\mathcal{T} \in \mathcal{F}}\ disc_{\mathcal{D},S,\mathcal{T}}$
	\STATE $\Lambda \gets$ {\sc Score\_Leaves($\mathcal{T}^\dagger,\mathcal{D},S$)} 
	\IF{$\Lambda=\emptyset$}
	    \STATE{$\mathcal{T}.flipped \gets true$}
	\ELSE
    	\STATE $\lambda^\dagger \gets  \argmax_{\lambda \in \Lambda}\ \lambda.score$
        \STATE{$\lambda^\dagger.flipped = true$}
        \STATE{$\lambda^\dagger.pred = 1 - \lambda.pred$}
        \STATE{$accu_{ \mathcal{F}}\gets  \frac{ | \mathcal{D}_{y= 1} \wedge \mathcal{F}(\bm{x})=1  |~+~ |\mathcal{D}_{y= 0 }\wedge \mathcal{F}(\bm{x})=0| }{|\mathcal{D}|} $}
      \STATE{$\Delta accu_{\mathcal{F}} \gets accu_{ \mathcal{F}}^* - accu_{ \mathcal{F}}$}
    \ENDIF
	\ENDWHILE
	\RETURN $\mathcal{F}$

	\end{algorithmic}
\end{algorithm}

\begin{table*}[ht]
\caption{Comparison of accuracy reduction and discrimination decrease on Adult dataset with respect to baseline accuracy of 0.85 and discrimination 0.2. 
Along with $\Delta$Accu and $\Delta$Disc, we also report (within parentheses) the final accuracy and discrimination values obtained.
\label{tab:adult}}
\center
\begin{tabular}{ccc|ccc|ccc|ccc|ccc|ccc|ccc|}
\cline{4-15}
& & & \multicolumn{2}{ c|| }{DFRF} & \multicolumn{2}{ c|| }{EOP} & \multicolumn{2}{ c|| }{EiFFFeL-TF} & \multicolumn{2}{ c|| }{EiFFFeL-LF}& \multicolumn{2}{ c||}{EiFFFeL-TF$^{\star}$} & \multicolumn{2}{ c|| }{EiFFFeL-LF$^{\star}$} \\ 
\cline{4-15}

& & & \multicolumn{1}{ c| }{$\Delta$Accu$\downarrow$} & \multicolumn{1}{ c|| }{$\Delta$Disc $\uparrow$}  & \multicolumn{1}{ c| }{$\Delta$Accu$\downarrow$} & \multicolumn{1}{ c|| }{$\Delta$Disc$\uparrow$} &
\multicolumn{1}{ c|}{$\Delta$Accu$\downarrow$} & \multicolumn{1}{ c||}{$\Delta$Disc$\uparrow$} & \multicolumn{1}{ c| }{$\Delta$Accu$\downarrow$} & \multicolumn{1}{ c|| }{$\Delta$Disc$\uparrow$} & \multicolumn{1}{ c| }{$\Delta$Accu$\downarrow$} & \multicolumn{1}{ c|| }{$\Delta$Disc$\uparrow$} & 
\multicolumn{1}{ c||}{$\Delta$Accu$\downarrow$} & \multicolumn{1}{ c|| }{$\Delta$Disc$\uparrow$}  \\ 
\toprule 

\multicolumn{1}{ |c}{\multirow{4}{*}{Adult} } & 
\multicolumn{1}{ |c  }{\multirow{4}{*}{\rotatebox[origin=c]{0}{$\epsilon$}}} &
\multicolumn{1}{|c|}{0.01} & 
\multicolumn{1}{c|}{{7(0.78)}} & \multicolumn{1}{c||}{{18(0.02)}}  &

\multicolumn{1}{ c|  }{\multirow{4}{*}{\rotatebox[origin=c]{0}{2(0.83)}}} &
\multicolumn{1}{ c||  }{\multirow{4}{*}{\rotatebox[origin=c]{0}{7(0.13)}}} &

\multicolumn{1}{c|}{4(0.81)} & \multicolumn{1}{c||}{19(0.01)}  & 
\multicolumn{1}{c|}{{4(0.81)}} & \multicolumn{1}{c||}{\color{blue}{\textbf{20(0)}}} 
& \multicolumn{1}{c|}{{6(0.79)}} & \multicolumn{1}{c||}{15(0.05)} 
& \multicolumn{1}{c|}{{3(0.82)}} & \multicolumn{1}{c||}{17(0.03)} 

\\
\cline{3-5}
\cline{8-15}
\multicolumn{1}{|c}{} & \multicolumn{1}{|c}{} & 
\multicolumn{1}{|c|}{0.05} & 
\multicolumn{1}{c|}{3(0.82)} & \multicolumn{1}{c||}{13(0.07)}  &

\multicolumn{1}{ c|  }{\multirow{4}{*}{\rotatebox[origin=c]{0}{}}} &
\multicolumn{1}{ c || }{\multirow{4}{*}{\rotatebox[origin=c]{0}{}}} &

\multicolumn{1}{c|}{3(0.82)} &
\multicolumn{1}{c||}{\color{blue}{\textbf{16(0.04)}}}  & \multicolumn{1}{c|}{{2(0.83)}} &

\multicolumn{1}{c||}{{15(0.05)}} 
& \multicolumn{1}{c|}{{6(0.79)}} & \multicolumn{1}{c||}{16(0.04)} 
& \multicolumn{1}{c|}{{3(0.82)}} & \multicolumn{1}{c||}{\color{blue}{\textbf{16(0.04)}}} 
\\
\cline{3-5}
\cline{8-15}
\multicolumn{1}{|c}{} & \multicolumn{1}{|c}{} & 
\multicolumn{1}{|c|}{0.10} & 
\multicolumn{1}{c|}{4(0.81)} & \multicolumn{1}{c||}{15(0.05)}  & 

\multicolumn{1}{ c|  }{\multirow{4}{*}{\rotatebox[origin=c]{0}{}}} &
\multicolumn{1}{ c||  }{\multirow{4}{*}{\rotatebox[origin=c]{0}{}}} &

\multicolumn{1}{c|}{2(0.83)} & \multicolumn{1}{c||}{12(0.08)}  & 
\multicolumn{1}{c|}{{1(0.84)}} & \multicolumn{1}{c||}{{\color{blue}{\textbf{12(0.08)}}}}
& \multicolumn{1}{c|}{{1(0.84)}} & \multicolumn{1}{c||}{\color{blue}{\textbf{12(0.08)}}} 
& \multicolumn{1}{c|}{{2(0.83)}} & \multicolumn{1}{c||}{10(0.1)} 
\\

\cline{3-5}
\cline{8-15}
\multicolumn{1}{|c}{} & \multicolumn{1}{|c}{} & 
\multicolumn{1}{|c|}{0.15} & 
\multicolumn{1}{c|}{2(0.83)} & \multicolumn{1}{c||}{10(0.1)}  &

\multicolumn{1}{ c|  }{\multirow{4}{*}{\rotatebox[origin=c]{0}{}}} &
\multicolumn{1}{ c||  }{\multirow{4}{*}{\rotatebox[origin=c]{0}{}}} &

\multicolumn{1}{c|}{0(0.85)} & \multicolumn{1}{c||}{8(0.12)}  & \multicolumn{1}{c|}{{0(0.85)}} & \multicolumn{1}{c||}{{ \color{blue}{\textbf{9(0.11)}}}}  
& \multicolumn{1}{c|}{{0(0.85)}} & \multicolumn{1}{c||}{7(0.13)} 
& \multicolumn{1}{c|}{{0(0.85)}} & \multicolumn{1}{c||}{7(0.13)} 
\\
\bottomrule

\end{tabular}

\end{table*}

\begin{table*}[ht]
\caption{Comparison of accuracy reduction and discrimination decrease on Bank dataset with respect to baseline accuracy of 0.82 and discrimination 0.18.  Along with $\Delta$Accu and $\Delta$Disc, we also report (within parentheses) the final accuracy and discrimination values obtained.
\label{tab:bank}}
\center
\footnotesize
\begin{tabular}{ccc|ccc|ccc|ccc|ccc|ccc|ccc|}
\cline{4-15}
& & & \multicolumn{2}{ c|| }{DFRF} & \multicolumn{2}{ c|| }{EOP} & \multicolumn{2}{ c|| }{EiFFFeL-TF} & \multicolumn{2}{ c|| }{EiFFFeL-LF}& \multicolumn{2}{ c||}{EiFFFeL-TF$^{\star}$} & \multicolumn{2}{ c|| }{EiFFFeL-LF$^{\star}$} \\ 
\cline{4-15}

& & & \multicolumn{1}{ c| }{$\Delta$Accu$\downarrow$} & \multicolumn{1}{ c|| }{$\Delta$Disc $\uparrow$}  & \multicolumn{1}{ c| }{$\Delta$Accu$\downarrow$} & \multicolumn{1}{ c|| }{$\Delta$Disc$\uparrow$} &
\multicolumn{1}{ c| }{$\Delta$Accu$\downarrow$} & \multicolumn{1}{ c|| }{$\Delta$Disc$\uparrow$} & \multicolumn{1}{ c| }{$\Delta$Accu$\downarrow$} & \multicolumn{1}{ c|| }{$\Delta$Disc$\uparrow$} & \multicolumn{1}{ c| }{$\Delta$Accu$\downarrow$} & \multicolumn{1}{ c|| }{$\Delta$Disc$\uparrow$} & 
\multicolumn{1}{ c||}{$\Delta$Accu$\downarrow$} & \multicolumn{1}{ c|| }{$\Delta$Disc$\uparrow$}  \\ 
\toprule 
\multicolumn{1}{ |c}{\multirow{4}{*}{Bank} } & 
\multicolumn{1}{ |c  }{\multirow{4}{*}{\rotatebox[origin=c]{0}{$\epsilon$}}} &
\multicolumn{1}{|c|}{0.01} & 
\multicolumn{1}{c|}{{9(0.73)}} & \multicolumn{1}{c||}{{13(0.05)}}  &

\multicolumn{1}{ c|  }{\multirow{4}{*}{\rotatebox[origin=c]{0}{0(0.82)}}} &
\multicolumn{1}{ c||  }{\multirow{4}{*}{\rotatebox[origin=c]{0}{14(0.04)}}} &

\multicolumn{1}{c|}{7(0.75)} & \multicolumn{1}{c||}{17(0.01)}  & \multicolumn{1}{c|}{{10(0.72)}} & \multicolumn{1}{c||}{15(0.03)} 
& \multicolumn{1}{c|}{{8(0.74)}} & \multicolumn{1}{c||}{14(0.04)} 
& \multicolumn{1}{c|}{{5(0.77)}} & \multicolumn{1}{c||}{10(0.08)} 
\\
\cline{3-5}
\cline{8-15}
\multicolumn{1}{|c}{} & \multicolumn{1}{|c}{} & 
\multicolumn{1}{|c|}{0.05} & 
\multicolumn{1}{c|}{4(0.78)} & \multicolumn{1}{c||}{11(0.07)}  & 

\multicolumn{1}{ c|  }{\multirow{4}{*}{\rotatebox[origin=c]{0}{}}} &
\multicolumn{1}{ c||  }{\multirow{4}{*}{\rotatebox[origin=c]{0}{}}} &

\multicolumn{1}{c|}{3(0.79)} & \multicolumn{1}{c||}{13(0.05)}  & \multicolumn{1}{c|}{{8(0.74)}} & \multicolumn{1}{c||}{{14(0.04)}} 
& \multicolumn{1}{c|}{{8(0.74)}} & \multicolumn{1}{c||}{13(0.05)} 
& \multicolumn{1}{c|}{{5(0.77)}} & \multicolumn{1}{c||}{13(0.05)} 
\\
\cline{3-5}
\cline{8-15}
\multicolumn{1}{|c}{} & \multicolumn{1}{|c}{} & 
\multicolumn{1}{|c|}{0.10} & 
\multicolumn{1}{c|}{4(0.78)} & \multicolumn{1}{c||}{6(0.12)}  &

\multicolumn{1}{ c|  }{\multirow{4}{*}{\rotatebox[origin=c]{0}{}}} &
\multicolumn{1}{ c||  }{\multirow{4}{*}{\rotatebox[origin=c]{0}{}}} &

\multicolumn{1}{c|}{2(0.80)} & \multicolumn{1}{c||}{10(0.08)}  & \multicolumn{1}{c|}{{1(0.81)}} & \multicolumn{1}{c||}{{7(0.11)}}
& \multicolumn{1}{c|}{{7(0.75)}} & \multicolumn{1}{c||}{9(0.09)} 
& \multicolumn{1}{c|}{{4(0.78)}} & \multicolumn{1}{c||}{8(0.10)} 
\\

\cline{3-5}
\cline{8-15}
\multicolumn{1}{|c}{} & \multicolumn{1}{|c}{} & 
\multicolumn{1}{|c|}{0.15} & 
\multicolumn{1}{c|}{4(0.78)} & \multicolumn{1}{c||}{9(0.09)}  &

\multicolumn{1}{ c|  }{\multirow{4}{*}{\rotatebox[origin=c]{0}{}}} &
\multicolumn{1}{ c||  }{\multirow{4}{*}{\rotatebox[origin=c]{0}{}}} &

\multicolumn{1}{c|}{0(0.82)} & \multicolumn{1}{c||}{4(0.14)}  & \multicolumn{1}{c|}{{0(0.82)}} & \multicolumn{1}{c||}{{ 4(0.14)}}  
& \multicolumn{1}{c|}{{6(0.76)}} & \multicolumn{1}{c||}{5(0.13)} 
& \multicolumn{1}{c|}{{2(0.80)}} & \multicolumn{1}{c||}{5(0.13)} 
\\
\bottomrule

\end{tabular}

\end{table*}


\begin{table*}[ht]
\caption{Comparison of accuracy reduction and discrimination decrease on Compas dataset with respect to baseline accuracy of 0.69 and discrimination 0.3.  
Along with $\Delta$Accu and $\Delta$Disc, we also report (within parentheses) the final accuracy and discrimination values obtained.
\label{tab:compas}}
\center
\footnotesize
\begin{tabular}{ccc|ccc|ccc|ccc|ccc|ccc|ccc|}
\cline{4-15}
& & & \multicolumn{2}{ c|| }{DFRF} & \multicolumn{2}{ c|| }{EOP} & \multicolumn{2}{ c|| }{EiFFFeL-TF} & \multicolumn{2}{ c|| }{EiFFFeL-LF}& \multicolumn{2}{ c||}{EiFFFeL-TF$^{\star}$} & \multicolumn{2}{ c|| }{EiFFFeL-LF$^{\star}$} \\ 
\cline{4-15}

& & & \multicolumn{1}{ c| }{$\Delta$Accu$\downarrow$} & \multicolumn{1}{ c|| }{$\Delta$Disc $\uparrow$}  & \multicolumn{1}{ c| }{$\Delta$Accu$\downarrow$} & \multicolumn{1}{ c|| }{$\Delta$Disc$\uparrow$} &
\multicolumn{1}{ c| }{$\Delta$Accu$\downarrow$} & \multicolumn{1}{ c|| }{$\Delta$Disc$\uparrow$} & \multicolumn{1}{ c| }{$\Delta$Accu$\downarrow$} & \multicolumn{1}{ c|| }{$\Delta$Disc$\uparrow$} & \multicolumn{1}{ c| }{$\Delta$Accu$\downarrow$} & \multicolumn{1}{ c|| }{$\Delta$Disc$\uparrow$} & 
\multicolumn{1}{ c||}{$\Delta$Accu$\downarrow$} & \multicolumn{1}{ c|| }{$\Delta$Disc$\uparrow$}  \\ 
\toprule 
\multicolumn{1}{ |c}{\multirow{4}{*}{COMPAS} } & 
\multicolumn{1}{ |c  }{\multirow{4}{*}{\rotatebox[origin=c]{0}{$\epsilon$}}} &
\multicolumn{1}{|c|}{0.01} & 
\multicolumn{1}{c|}{{11(0.58)}} & \multicolumn{1}{c||}{{28(0.02)}}  &

\multicolumn{1}{ c|  }{\multirow{4}{*}{\rotatebox[origin=c]{0}{4(0.65)}}} &
\multicolumn{1}{ c||  }{\multirow{4}{*}{\rotatebox[origin=c]{0}{5(0.25)}}} &

\multicolumn{1}{c|}{25(0.44)} & \multicolumn{1}{c||}{29(0.01)}  & \multicolumn{1}{c|}{{5(0.64)}} & \multicolumn{1}{c||}{26(0.04)} 
& \multicolumn{1}{c|}{{9(0.60)}} & \multicolumn{1}{c||}{29(0.01)} 
& \multicolumn{1}{c|}{{1(0.68)}} & \multicolumn{1}{c||}{7(0.23)} 
\\
\cline{3-5}
\cline{8-15}
\multicolumn{1}{|c}{} & \multicolumn{1}{|c}{} & 
\multicolumn{1}{|c|}{0.05} & 
\multicolumn{1}{c|}{5(0.64)} & \multicolumn{1}{c||}{13(0.17)}  &

\multicolumn{1}{ c|  }{\multirow{4}{*}{\rotatebox[origin=c]{0}{}}} &
\multicolumn{1}{ c||  }{\multirow{4}{*}{\rotatebox[origin=c]{0}{}}} &

\multicolumn{1}{c|}{12(0.57)} & \multicolumn{1}{c||}{28(0.02)}  & \multicolumn{1}{c|}{{5(0.64)}} & \multicolumn{1}{c||}{{22(0.08)}} 
& \multicolumn{1}{c|}{{9(0.60)}} & \multicolumn{1}{c||}{28(0.02)} 
& \multicolumn{1}{c|}{{1(0.68)}} & \multicolumn{1}{c||}{7(0.23)} 
\\
\cline{3-5}
\cline{8-15}
\multicolumn{1}{|c}{} & \multicolumn{1}{|c}{} & 
\multicolumn{1}{|c|}{0.10} & 
\multicolumn{1}{c|}{4(0.65)} & \multicolumn{1}{c||}{7(0.23)}  & 

\multicolumn{1}{ c|  }{\multirow{4}{*}{\rotatebox[origin=c]{0}{}}} &
\multicolumn{1}{ c||  }{\multirow{4}{*}{\rotatebox[origin=c]{0}{}}} &

\multicolumn{1}{c|}{7(0.62)} & \multicolumn{1}{c||}{21(0.09)}  & \multicolumn{1}{c|}{{5(0.64)}} & \multicolumn{1}{c||}{{21(0.09)}}
& \multicolumn{1}{c|}{{1(0.68)}} & \multicolumn{1}{c||}{21(0.09)} 
& \multicolumn{1}{c|}{{1(0.68)}} & \multicolumn{1}{c||}{7(0.23)} 
\\

\cline{3-5}
\cline{8-15}
\multicolumn{1}{|c}{} & \multicolumn{1}{|c}{} & 
\multicolumn{1}{|c|}{0.15} & 
\multicolumn{1}{c|}{2(0.67)} & \multicolumn{1}{c||}{6(0.24)}  &

\multicolumn{1}{ c|  }{\multirow{4}{*}{\rotatebox[origin=c]{0}{}}} &
\multicolumn{1}{ c||  }{\multirow{4}{*}{\rotatebox[origin=c]{0}{}}} &

\multicolumn{1}{c|}{1(0.68)} & \multicolumn{1}{c||}{19(0.11)}  & \multicolumn{1}{c|}{{2(0.67)}} & \multicolumn{1}{c||}{{ 15(0.15)}}  
& \multicolumn{1}{c|}{{0(0.69)}} & \multicolumn{1}{c||}{16(0.14)} 
& \multicolumn{1}{c|}{{1(0.68)}} & \multicolumn{1}{c||}{7(0.23)} 
\\
\bottomrule

\end{tabular}

\end{table*}

\section{Experimental Evaluation}

\subsection{Datasets.} 
We use datasets publicly available, widely used in fairness literature,  concerning binary classification. We pre-process them using one-hot encoding for categorical features, binary encoding of sensitive feature, and removing of instances containing missing values. Moreover, we use an 80/20 training/test split.

\begin{itemize}
    \item \textit{Adult}: The Adult UCI income dataset \cite{Dua:2019} contains 14 demographic attributes of more than 45,000 individuals, together with class labels which states whether their income is higher than \$50K or not. As sensitive attribute, we use the \textit{gender} encoded as a binary attribute 1/0 for male/female respectively.
    \item \textit{COMPAS}: The COMPAS dataset  \cite{angwin2016machine} contains data collected on the use of the COMPAS (Correctional Offender Management Profiling for Alternative Sanctions)  risk assessment tool. It  contains 13 attributes of more than 7,000 convicted criminals, with class labels that state whether or not the individual reoffend within 2 years of her/his most recent crime. We use \textit{race} as sensitive attribute encoded as a binary attribute 1/0 for Others/African-American, respectively. 
    \item \textit{Bank}: Bank marketing dataset \cite{moro2014data} contains 16 features about 45,211 clients of direct marketing campaigns of a Portuguese banking institution. The goal is to predict whether the client will subscribe or not to a term deposit. We consider the \textit{age} as sensitive attribute, encoded as a binary attribute 1/0, indicating whether the client’s age is $\geq$25 or $<$25,  respectively.
\end{itemize}

\begin{figure*} [ht]
\centering

\includegraphics[width=0.7\textwidth]{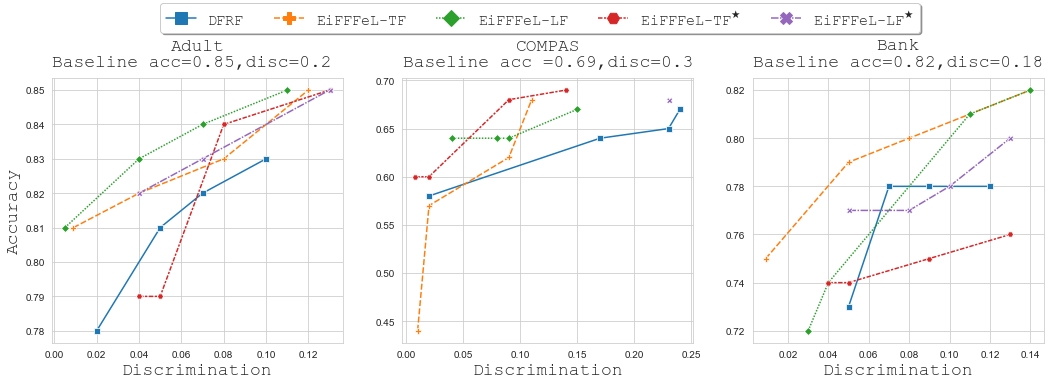}
\caption{ Accuracy vs. discrimination scores after relabeling for constraints $\epsilon=0.01,0.05,0.1,0.15$. }
\label{fig:plot1}
\end{figure*}

\begin{figure*}[ht]
\centering
\includegraphics[width=0.7\textwidth]{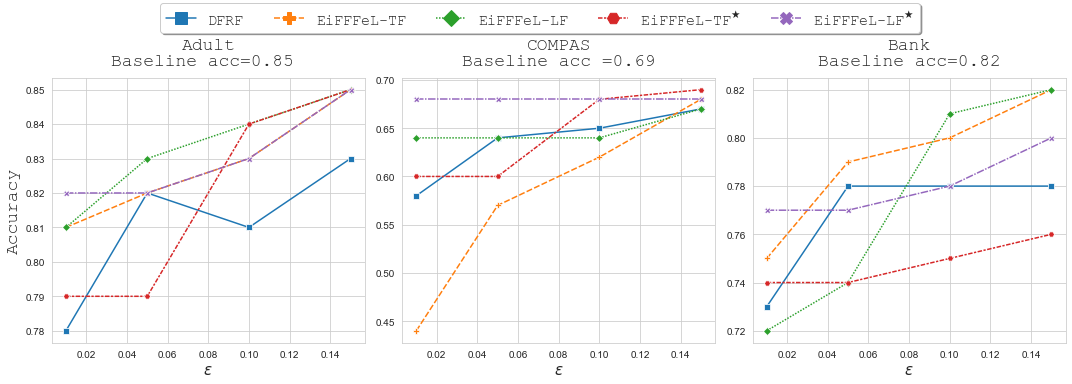}
\caption{Accuracy of the model as a function the $\epsilon$ constraint.}
\label{fig:plot2}
\end{figure*}

\begin{figure*}[ht]
\centering
\includegraphics[width=0.7\textwidth]{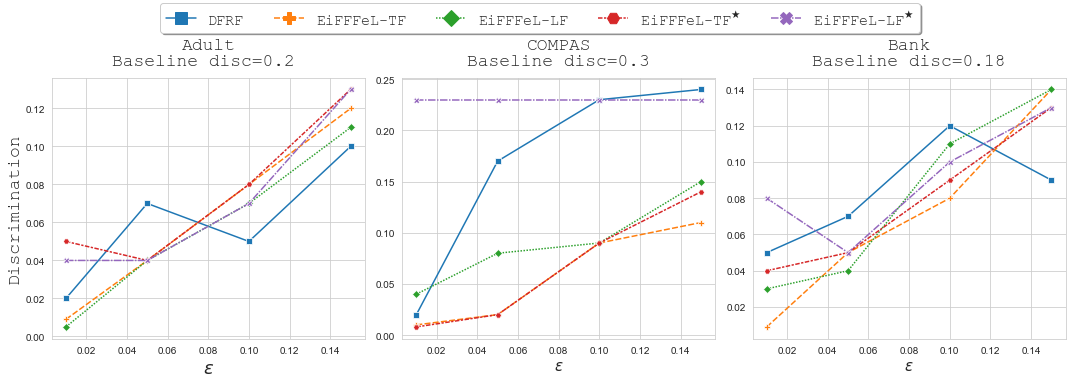}
\caption{Discrimination scores as a function of the $\epsilon$   constraint.}
\label{fig:plot3}
\end{figure*}

\subsection{Experimental Setup.} 
We apply our proposed EiFFFeL algorithm over a Random Forest classifier  with/without the fair splitting of nodes for individual base trees, and evaluate the performance of the algorithms in terms of model accuracy and discrimination over the three datasets mentioned above. 

We compare our results against a DFRF classifier (\textit{Distributed fair random forest}) \cite{fantin2020distributed}, which only includes fair decision trees within the forest.  The setting of hyper-parameters of DFRF are the same as the one described in the original work. We use fair split and sensitive feature  as hyper-parameters, along with
tree number and maximum tree depth. 
Additionally, 
we also compare our results against EOP (\textit{Equalized Odds Post-processing}) \cite{hardt2016equality,pleiss2017fairness}, a random forest classifier  with the same number of base estimators and maximum depth as ours. After training and achieving the desired equalized odd we score the discrimination in the same approach we used for our experiments.

In conclusion, the comparisons of accuracy and discrimination values are among the following methods:

\begin{itemize}
\item DFRF \cite{fantin2020distributed},\footnote{\url{https://github.com/pjlake98/Distributed-Fair-Random-Forest}} which adds base trees to the forest only if they are fair; 

\item EOP \cite{hardt2016equality,pleiss2017fairness},\footnote{\url{https://github.com/Trusted-AI/AIF360/blob/master/aif360/algorithms/postprocessing/calibrated_eq_odds_postprocessing.py}} which adopts a post-processing method based on achieving equalized odds requiring the privileged and unprivileged groups to have the same false negative rate and same false positive rate;

\item our implementations of EiFFFeL-TF and  EiFFFeL-LF algorithms, whose post-processing is applied to a plain Random Forest of trees; 

\item the same post-processing techniques of EiFFFeL-TF and  EiFFFeL-LF applied on top of a random forest with discrimination aware base trees \cite{kamiran2010discrimination}. These versions are denoted by  EiFFFeL-TF$^{\star}$ and EiFFFeL-LF$^{\star}$.
\end{itemize}

Finally, the \textit{baseline accuracy} and \textit{discrimination} used to compare the various methods are the ones obtained by a plain Random Forest of trees, trained on the three datasets through the scikit-learn algorithm Random Forest Classifier\footnote{\url{https://scikit-learn.org/stable/modules/generated/sklearn.ensemble.RandomForestClassifier.html}}. The various EiFFFeL methods are applied to the same baseline Random Forest.

\subsection{Results.}

Tables \ref{tab:adult}), \ref{tab:bank}), and \ref{tab:compas}) compare the decreases in accuracy and discrimination, obtained by the different algorithms, on the three datasets with respect to the baseline results obtained by plain Random Forest models.

Recall that increasing $\epsilon$, we reduce the space for improving discrimination, and as a side effect, we preserves the baseline accuracy. Indeed, in these experiments the accuracy constraint $\alpha$ was set to 1, so that there are no limits in the possible accuracy reduction $\Delta$Accu. This  allows us to compare our methods against DFRF and EOP, which do not have this $\alpha$  constraint. Indeed, EOP is completely parameter free, and does not support neither $\alpha$ nor $\epsilon$.

In more details, Tables \ref{tab:adult}), \ref{tab:bank}), and \ref{tab:compas}) report, for different values of $\epsilon$ in the set $\{0.01, 0.05, 0.10, 0.15\}$, the $\Delta$Accu and $\Delta$Disc values obtained by the different algorithms, where $\Delta$Accu and $\Delta$Disc indicate the \textit{absolute difference} in accuracy and discrimination w.r.t. the baselines.
Indeed, we express these $\Delta$ absolute differences in points/hundredths (each point corresponds to $1/100$).
Note that while greater values of $\Delta$Disc are better, greater values of $\Delta$Accu are worse, so a trade-off is needed. In addition, besides the absolute $\Delta$ values, we also report (within parentheses) the final values for accuracy and discrimination score obtained by the various techniques.

For example, for the \textit{Adult} dataset (Table \ref{tab:adult}) and $\epsilon=0.01$, EiFFFeL-TF can reach a very low discrimination score of $0.01$, by only losing $4$ points in accuracy (from $0.85$ of the baseline to $0.81$). In comparison, the best results we can obtain with DFRF in terms of discrimination is a score of $0.02$, by losing 7 points in accuracy (from $0.85$ of the baseline to $0.78$). Overall, our algorithms are capable of reducing discrimination better than DFRF while maintaining the same accuracy. Also EOP does not work well, as the best discrimination score is only $0.13$, by losing 2 points in accuracy. In addition, using $\epsilon=0.15$ for EiFFFeL-TF and EiFFFeL-LF (also EiFFFeL-TF$^{\star}$ and EiFFFeL-LF$^{\star}$), we can decrease the baseline discrimination of about $7-9$ points,  by keeping the same accuracy of the baseline.

Results for the Bank dataset (Table \ref{tab:bank}) shows that EiFFFeL-TF can reach for $\epsilon=0.01$ the desired discrimination score, but losing 7 points in accuracy (from 0.82 to 0.75), whereas DFRF has worse discrimination score of 0.05 and a worse accuracy of 0.73. EOP does not lose any accuracy for lowering the discrimination score by 14 points to 0.04.  

Finally, considering the results obtained for the COMPAS dataset (Table \ref{tab:compas}), we observe in some cases DFRF works pretty well, but always one of our algorithms gets better results. For example, for $\epsilon=0.01$, the best discrimination score of $0.01$ is obtained by EiFFFeL-TF$^\star$, by only losing 9 points in accuracy, against the 11 points lost by DFRF with a discrimination score of $0.02$.

\medskip
 
Figures~\ref{fig:plot1}, \ref{fig:plot2}, and \ref{fig:plot3} report the same data of the above tables, where we varied the discrimination constraint $\epsilon=\{0.01, 0.05, 0.1, 0.5\}$, with no constraints on accuracy. The results obtained by EOP are not plotted, as its results are always worse than the competitors and do not vary with $\epsilon$.

Specifically, Figure \ref{fig:plot1} reports results for the three datasets, and aims at showing the tradeoff 
of accuracy vs. discrimination when we vary $\epsilon$. Recall that we are interested in achieving low discrimination and high accuracy, and thus the best tradeoff corresponds to points of curves falling in the top-left quadrant.  

First, we highlight that  DRFR performs poorly on most settings compared to the proposed EIFFFeL variants. 
On the Adult dataset, EIFFFeL-LF dominates the other algorithms for all values of $\epsilon$ and achieves the desired or better discrimination with the largest accuracy. To appreciate the strict relationships between of the setting of $\epsilon$ and the discrimination/accuracy obtained, the reader can refer to the other two Figures \ref{fig:plot1} and \ref{fig:plot2}. 

Returning to Figure \ref{fig:plot1}, the COMPAS EIFFFeL-LF provides the best performance together with EIFFFeL-TF$^{\star}$. This is the only dataset where EIFFFeL-TF$^{\star}$ provides interesting performance, and thus the discrimination aware splitting at training time provides some benefits. We also highlight that when using $\epsilon=0.15$ (see Figure \ref{fig:plot3}) the algorithm DFRF only gets a discrimination score of 0.25. Note that EIFFFeL-LF$^{\star}$ is not able to provide better performance when varying $\epsilon$, thus resulting in a constant curve. 

Finally, on the Bank Dataset, EIFFFeL-TF and EIFFFeL-LF achieve the best results, with an advantage for EIFFFeL-TF for smaller values of $\epsilon$. Finally, the results show how we can obtain the desired discrimination degree  with a limited drop in accuracy. Overall, the proposed EIFFFEL algorithm outperforms the competitor DFRF, and, on average, it is advisable to avoid the discrimination aware node splitting. We believe that working only at post-processing allows us to exploit a richer set of trees grown, by exploring a larger and unconstrained search space.

\begin{table*}

\caption{Accuracy and discrimination scores on the Adult dataset for  $\epsilon=0.01$ and $\alpha=0.01,0.02,0.03,0.05$. The baseline accuracy and discrimination score are $0.85$ and $0.2$, respectively. 
\label{tab:alpha}}
\center
\footnotesize
\begin{tabular}{cccc|ccc|ccc|ccc|}
\cline{5-12}
& & & & \multicolumn{2}{ c|| }{EiFFFeL-TF} & \multicolumn{2}{ c|| }{EiFFFeL-LF}& \multicolumn{2}{ c||}{EiFFFeL-TF$^{\star}$} & \multicolumn{2}{ c|| }{EiFFFeL-LF$^{\star}$} \\ 
\cline{5-12}

& & & & \multicolumn{1}{ c| }{Accu} & \multicolumn{1}{ c|| }{Disc} & \multicolumn{1}{ c| }{Accu} & \multicolumn{1}{ c|| }{Disc} & \multicolumn{1}{ c| }{Accu} & \multicolumn{1}{ c|| }{Disc} & \multicolumn{1}{ c||}{Accu} & \multicolumn{1}{ c|| }{Disc}  \\ 
\toprule 
\multicolumn{1}{ |c}{\multirow{4}{*}{Adult} } & 
\multicolumn{1}{ |c  }{\multirow{4}{*}{\rotatebox[origin=c]{0}{$\epsilon$=0.01}}} &
\multicolumn{1}{ |c  }{\multirow{4}{*}{\rotatebox[origin=c]{0}{$\alpha$}}} &

\multicolumn{1}{|c|}{0.01} & 
\multicolumn{1}{c|}{0.83} & \multicolumn{1}{c||}{0.09}  & 
\multicolumn{1}{c|}{{0.84}} & \multicolumn{1}{c||}{\bf0.08} 
& \multicolumn{1}{c|}{{0.84}} & \multicolumn{1}{c||}{0.10} 
& \multicolumn{1}{c|}{{0.84}} & \multicolumn{1}{c||}{0.11} 
\\ 
\cline{4-12}
\multicolumn{1}{|c}{} & 
\multicolumn{1}{|c}{} &
\multicolumn{1}{|c}{} & 

\multicolumn{1}{|c|}{0.02} & 
\multicolumn{1}{c|}{0.83} & \multicolumn{1}{c||}{0.09}  &
\multicolumn{1}{c|}{{0.83}} & \multicolumn{1}{c||}{{\bf0.06}} 
& \multicolumn{1}{c|}{{0.83}} & \multicolumn{1}{c||}{0.10} 
& \multicolumn{1}{c|}{{0.83}} & \multicolumn{1}{c||}{0.07} 
\\
\cline{4-12}
\multicolumn{1}{|c}{} & 
\multicolumn{1}{|c}{} &
\multicolumn{1}{|c}{} & 

\multicolumn{1}{|c|}{0.03} & 
\multicolumn{1}{c|}{0.82} & \multicolumn{1}{c||}{\bf0.04}  & 
\multicolumn{1}{c|}{{0.82}} & \multicolumn{1}{c||}{{0.05}}
& \multicolumn{1}{c|}{{0.82}} & \multicolumn{1}{c||}{0.07}
& \multicolumn{1}{c|}{{0.82}} & \multicolumn{1}{c||}{\bf0.04} 
 
\\

\cline{4-12}
\multicolumn{1}{|c}{} & 
\multicolumn{1}{|c}{} &
\multicolumn{1}{|c}{} & 
 
\multicolumn{1}{|c|}{0.05} & 
\multicolumn{1}{c|}{0.81} & \multicolumn{1}{c||}{0.01}  &
\multicolumn{1}{c|}{{0.81}} & \multicolumn{1}{c||}{{\bf 0.00}}  
& \multicolumn{1}{c|}{{0.80}} & \multicolumn{1}{c||}{0.08} 
& \multicolumn{1}{c|}{{0.82}} & \multicolumn{1}{c||}{0.03} 
\\
\bottomrule

\end{tabular}

\end{table*}

The effect of varying the discrimination constraint $\epsilon$ without constraining accuracy can be observed in Figure \ref{fig:plot2}, where we discover that lower discrimination is achieved with large accuracy reduction. This is due to the fact that a small discrimination threshold allows our flipping strategies to force the change of many leaves, thus changing more the classification decision regions, with a final lower accuracy. 
However our approach of selecting potential leaves to relabel seems better  than training random forest with only fair trees. 
In addition, training and then rejecting trees (because they are not fair) makes longer the training of the forest,   
particularly when we fail often in finding fair trees.

Finally, Figure~\ref{fig:plot3} contrasts the discrimination measured  on the test set against the desired discrimination constraint $\epsilon$. Clearly, the twos do not always match. In particular, DFRF has an unstable behaviour, meaning that filtering the tree to be added to the forest is not the best option. Conversely, EiFFFeL-TF and EiFFFeL-LF provide a much more stable behaviour.

\medskip

We also discuss the results of other experiments, aiming to evaluate the effects of different values for the $\alpha$ constraints. Note that only the EiFFFeL algorithms   support the $\alpha$ parameter, so we cannot reports any results for  the competitors DFRF and EOP. 
Specifically, Table \ref{tab:alpha} reports results relative to the Adult dataset, where,
for a fixed $\epsilon=0.01$,  we vary the $\alpha$ constraint over the expected accuracy, with values ranging in the set \{0.01,0.02,0.03,0.05\}. For each $\alpha$ value, we show in bold the best results in terms of discrimination score.
We observe that the accuracy constraint $\alpha$ has an indirect impact on the final discrimination score obtained.  Using EiFFFeL-LF with $\alpha=0.01$, the loss in accuracy is 1 point as expected, while the baseline discrimination score decreases by more than half (from $0.2$ to $0.08$). Furthermore, as the $\alpha$ value increases, discrimination score decreases further. With $\alpha=0.05$, EiFFFeL-LF is able to reduce by 4 points the final accuracy, by also achieving a discrimination score of 0,  thus showing the power of our method in achieving a very good trade-off between accuracy and discrimination.

\section{Conclusion}
In this work we deal with fairness in machine learning, and specifically in binary classifiers trained by a Random Forest algorithm. We are interested in group fairness, so as to  mitigate the effect of bias against specific groups, which may comes from biased training datasets or algorithm design.

We develop EiFFFeL, a novel post-process approach, which maintains good predictive performance of the trained model with a low discrimination score. Our approaches flips the label of selected leaf (or leaves) of base trees in a random forest by using two algorithms: $(i)$ an aggressive tree-based approach, which flips all candidate leaves of a tree, and $(ii)$ a leaf-based strategy which only flips the label of the most discriminative leaf of a tree. Both strategies are implemented by considering accuracy and discrimination constraints. Indeed, the constraints are used  to control the minimum accuracy decrease we can tolerate in order to achieve the desired discrimination value. 
In addition, we have tested the impact of incorporating discrimination aware node split strategies for base trees of the forest, by adding discrimination gain value in their node splitting criterion \cite{kamiran2010discrimination}.

By using three publicly available datasets, our experimental results show that effective non-discriminative models can be obtained, while keeping  a strict control over both accuracy and discrimination level.  Compared to   state-of-the-art methods, which adopt both in-process and post-process bias mitigation approaches, EiFFFeL resulted to produce the most accurate models that also exhibit the best levels of fairness. 

As part of the future work, we plan to extend our methods by studying the effect of multiple sensitive features in relation to discrimination and accuracy, by also extending our work to other tree ensemble learning methods.

\bibliographystyle{ACM-Reference-Format}
\bibliography{main} 

\end{document}